\pdfoutput=1

\documentclass[runningheads]{llncs}
\usepackage[T1]{fontenc}
\usepackage{graphicx,verbatim}
\usepackage[numbers,sort]{natbib}
\usepackage{booktabs} %
\usepackage{amsmath}
\usepackage{amssymb}  %
\usepackage{multirow}
\usepackage{xcolor}
\usepackage{marvosym}

\usepackage{amsmath}
\usepackage{overpic} %
\usepackage{enumitem} %
\usepackage{color}
\usepackage{colortbl} %
\usepackage{adjustbox} %
\usepackage{tabularx} %
\usepackage{soul} %
\usepackage{dblfloatfix}
\usepackage{wrapfig}
\usepackage[T1]{fontenc}
\usepackage{lipsum} 
\usepackage{graphicx}
\usepackage{float}
\usepackage{placeins}
\definecolor{forestgreen}{rgb}{0.13, 0.55, 0.13}
\usepackage[breaklinks,colorlinks,citecolor = forestgreen,hidelinks]{hyperref}

\usepackage[capitalise]{cleveref}

\definecolor{teal}{rgb}{0,.5,.5}
\definecolor{ligh_teal}{rgb}{0.51,.79,.77}
\definecolor{blue}{rgb}{0,0,1}
\definecolor{red}{rgb}{1,0,0}
\definecolor{green}{rgb}{0,1.,0}
\definecolor{orange}{rgb}{0.75, 0.4, 0}
\definecolor{turquoise}{cmyk}{0.65,0,0.1,0.3}
\definecolor{purple}{rgb}{0.65,0,0.65}
\definecolor{darkjunglegreen}{rgb}{0.1, 0.14, 0.13}
\definecolor{darkgreen}{rgb}{0.0, 0.2, 0.13}
\definecolor{darkred}{rgb}{0.6, 0.1, 0.05}
\definecolor{blueish}{rgb}{0.0, 0.3, .6}
\definecolor{light_gray}{rgb}{0.7, 0.7, .7}
\definecolor{pink}{rgb}{1, 0, 1}
\definecolor{greyblue}{rgb}{0.25, 0.25, 1}
\definecolor{magenta}{rgb}{1., 0., 1.}

\newcommand{\ourModel}{CrownFusion}
\newcommand{\ourRepresentation}{CrownImage}

\usepackage{blindtext}
\newcommand{\lorem}[1]{}

\renewcommand{\paragraph}[1]{\vspace{1em}\noindent\textbf{#1}.}
\newcommand{\mypara}[1]{\textbf{#1}.}

\begin{document}
\title{CrownFusion: 3D Dental Crown Generation Using Geometry Images and Latent Diffusion}
\titlerunning{CrownFusion}
\author{Johan Ziruo Ye\inst{1,2}$^{\star}$ \and
Xingguang Yan\inst{3}$^{\star}$~(\Letter) \and
Søren Hauberg\inst{1}  \and
 \\ Angel X. Chang\inst{3,4} \and
Hao Zhang\inst{3} \and
Peter Lempel Søndergaard\inst{2}}
\index{Ye, Johan Ziruo}
\index{Yan, Xingguang}
\index{Hauberg, Søren}
\index{Chang, Angel X.}
\index{Zhang, Hao}
\index{Søndergaard, Peter Lempel}
\authorrunning{J. Ye et al.}
\institute{Technical University of Denmark, Kongens Lyngby, Denmark \and
3Shape, Copenhagen, Denmark \and
Simon Fraser University, Vancouver, Canada \and
Canada CIFAR AI Chair at Amii, Edmonton, Canada \\
\email{qheldiv@gmail.com}}
\maketitle
{\renewcommand\thefootnote{$\star$}\footnotetext{Equal contribution.}}%
{\renewcommand\thefootnote{$\dagger$}\footnotetext{Code and data available at \url{https://github.com/JohanYe/CrownFusion}.}}
\begin{abstract}
We introduce \emph{CrownFusion}$^{\dagger}$, a latent diffusion model to generate 3D dental crown designs from \emph{geometry images}, a 2D-grid encoding of surface coordinates. Departing from traditional approaches that use point clouds, our 
``\emph{CrownImage}'' representation not only unlocks the full power of image-based generative architectures, but also preserves fine geometric detail by operating at far higher spatial resolutions than point clouds to alleviate smoothing artifacts in the 3D generation.
Conditioned on jointly learned embeddings of the surrounding dentition, our probabilistic diffusion model produces anatomically plausible crowns with morphological variations tailored to individual patient cases. 
To accommodate multiple valid designs, we propose an occlusal fit metric based on proximity to the antagonist teeth. We evaluate our method through quantitative experiments and qualitative studies involving expert testimonials.
We also release the \emph{FDI 16 Crown Dataset}, comprising 11,513 annotated intra-oral scans, to support future research in dental CAD.


\end{abstract}
\section{Introduction}
\label{sec:intro}

Computer-aided dental design software has streamlined the creation of artificial tooth replacements, allowing technicians to adapt templates to individual patients. Yet, crown design remains labor-intensive, requiring precise adaptation to each patient's unique dental morphology. A clinically acceptable crown must achieve adequate occlusal fit with the opposing dentition, 
maintain functional contact without hyperocclusion \cite{bronson_clinical_2005}, and reflect natural tooth morphology, including well-defined occlusal grooves that direct food flow during mastication. 
In practice however, there is no single correct crown for a given case. Multiple clinically acceptable designs may exist, shaped by the technician's training, experience, and aesthetic judgement \cite{tian_dual_2022, mccracken_laboratory_2020}. 
This motivates a generative approach to capture a \emph{space} of clinically acceptable solutions rather than predicting a single outcome, as well as developing a suitable metric for evaluation.\looseness=-1

In terms of shape representations, previous approaches to automated dental crown design face fundamental limitations. 
Point cloud-based methods \cite{yang_dcrownformer_2024, hosseinimanesh_mesh_2025, wei_vbcd_2025} frame crown design as a shape completion task, requiring a separate mesh extraction step that typically introduces surface artifacts. %
They tend to produce overly smooth outputs and erode clinically important details such as sharp fissures, due to the difficulty of placing points at high-curvature features and optimization using
the Chamfer Distance \citep{barrow_parametric_1977} (CD), which %
is insensitive to local geometric details.
Depth image-based methods \cite{hwang_learning_2018, yuan_personalized_2020, tian_dual_2022} are confined to projection from a single viewpoint, inherently discarding 3D information. 
For instance, the lingual surface of an antagonist tooth, which directly contacts the generated crown during occlusion, may be entirely occluded and absent from the depth rendering.
 
\begin{figure}[!t]
    \centering
    \includegraphics[width=\linewidth]{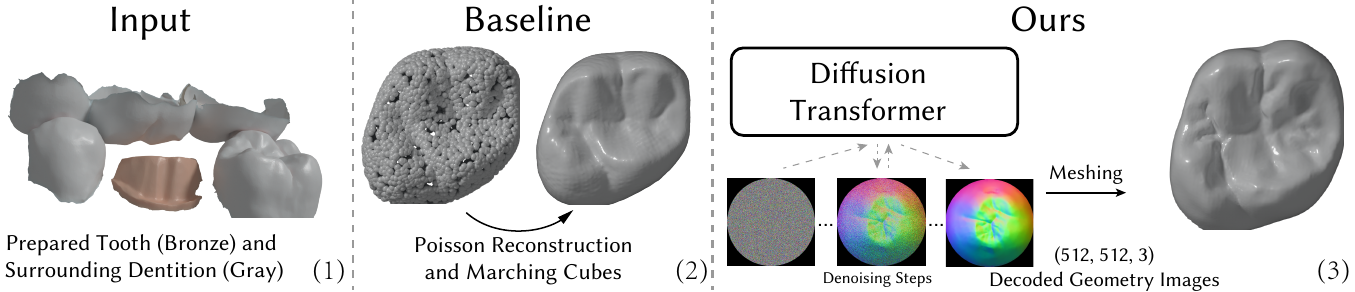}
    \caption{\textbf{Overview.} From left to right: (1) Input: prepared tooth (bronze) and surrounding dentition (gray) from an intraoral scan. (2) Baseline: point cloud-based methods with Poisson reconstruction %
    yield overly smooth geometry eroding occlusal detail. (3) Ours: a diffusion transformer generates a geometry image encoding of the crown surface, which is then meshed to recover the 3D crown, preserving fine occlusal features.
    }
    \label{fig:teaser}
\end{figure}

\begin{figure}[!t]
    \centering
    \includegraphics[width=\linewidth]{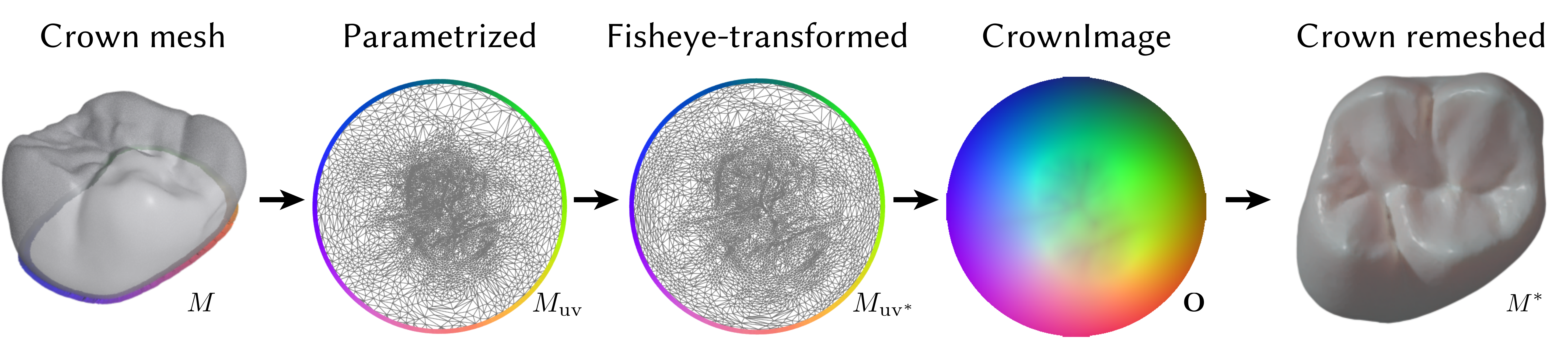}
    \caption{\ourRepresentation{} construction pipeline. $M$: crown mesh. $M_{\mathrm{uv}}$: parametrized mesh on the unit disk via boundary-fixed harmonic mapping. $M_{\mathrm{uv}*}$: fisheye-transformed UV mesh (inner disk expanded, outer annulus compressed). $\mathbf{O}$: rasterized geometry image encoding $(x,y,z)$ and $\alpha$. $M^*$: remeshed crown recovered by triangulating $\mathbf{O}$.
    }    \label{fig:geometry_image_representation_path}
\end{figure}

In this work, we address these limitations by generating 3D dental crowns through \emph{geometry images} \citep{gu_geometry_2002}, which encode a tooth's surface geometry into a 2D grid, i.e., a \emph{CrownImage}, with each entry storing an $[x, y, z]$ coordinate on the surface as color 
(\cref{fig:geometry_image_representation_path}). 
Such a 3D-to-2D mapping is facilitated by the reliable manifold topology of tooth meshes, ensuring a seamless parameterization onto the 2D grid. Unlike single-view depth renderings, geometry images retain full 3D geometric information, while supporting far higher spatial resolutions than point clouds to better preserve fine-grained and high-curvature teeth features. 

\ourRepresentation{}s, our image-based teeth representation, enables us to develop \emph{CrownFusion}, a conditional latent diffusion model for 3D crown generation. To our knowledge, this is both the first use of geometry images and the first latent diffusion model for this task. We combine the Scalable Interpolant Transformer (SiT) \citep{ma_sit_2024} and a FLUX VAE \citep{black_forest_labs_flux_2024}, with cross-attention conditioning on shape embeddings of the surrounding dentition (\cref{fig:teaser}). The generation of multiple morphologically varied crown candidates for a single patient offers clinicians a range of viable designs to explore. To complement standard CD evaluation, we propose an occlusal fit metric based on proximity to the antagonist teeth to better capture functional performance. Experiments show that \ourModel{} achieves competitive occlusal alignment with existing completion-based methods while producing substantially fewer intersections. Expert testimonials confirm that our method consistently produces more well-defined and natural-looking occlusal anatomy and was less prone to blunt or anatomically incoherent surfaces than baselines. Finally, we release the \emph{FDI 16 Crown Dataset} of 11,513 annotated intraoral scans.

\section{Related Work}

\mypara{Crown design via 3D completion} 
Recent methods frame dental crown generation as a 3D shape completion task. 
\textit{DCrownFormer} \cite{yang_dcrownformer_2024} and 
\textit{DCrownFormer+} \cite{yang_dcrownformer+_2025} encode antagonist and preparation tooth point clouds via transformer blocks and a morphology-aware cross-attention decoder. 
Similarly, \textit{DMC} \cite{hosseinimanesh_mesh_2025} designs a transformer-based encoder-decoder followed by FoldingNet \cite{yang_foldingnet_2018} to output a point cloud. Recently, \textit{VBCD} \cite{wei_vbcd_2025} voxelizes intraoral scans and generates coarse crowns via a 3D UNet followed by point cloud refinement.
Relying on Shape-as-Points \cite{peng_shape_2021} for final mesh extraction, these methods are all deterministic completion approaches with a single outcome per input. In contrast, \ourModel{} is fully generative and produces multiple designs based on the surrounding teeth.

\mypara{Crown design via depth image synthesis} Converting dental meshes to depth images is another way to enable the use of 2D deep learning methods. Hwang et al.~\cite{hwang_learning_2018} applied GANs \cite{goodfellow_generative_2014} to synthesize crown depth images, though results varied in anatomical correctness. Yuan et al.~\cite{yuan_personalized_2020} extended this with a conditional GAN incorporating neighboring teeth and an occlusal groove filter loss to improve fissure alignment. Tian et al.~\cite{tian_dual_2022} proposed a dual-GAN framework with global and local discriminators to enforce both contextual consistency and morphological detail. Like \ourModel{}, these methods are generative, but they lack the controllable generation afforded by diffusion models.
In addition, generating a complete crown represented by depth images is challenging as it may require generating multiple views that are consistent.

\mypara{Geometry images} Geometry images have primarily been explored for 3D objects with simple topology. \citet{sinha_deep_2016} first applied 2D deep learning to geometry images for shape classification, using authalic spherical parameterization to unwrap closed meshes. \citet{maron_convolutional_2017} extended this to toric objects with an additional dimension and stitching to handle topological mismatches. Recently, \citet{yan_object_2024} proposed a latent diffusion pipeline that segments objects into simple patches, maps each to geometry images, and stitches them during reconstruction. In contrast, intraoral scans are open surfaces, inherently compatible with geometry images without complex parameterization or stitching.%

\section{Approach}
Our core contribution is the introduction of \ourRepresentation{}s, geometry images~\cite{gu_geometry_2002} of dental crowns, as a data representation that enables the use of established 2D generative architectures. We train \ourModel{}, a latent diffusion model that generates \ourRepresentation{}s conditioned on the surrounding dentition, and contribute the FDI 16 Crown Dataset for training and evaluation. Next, we describe the parameterization, our diffusion pipeline, and the dataset.\looseness=-1

\mypara{\ourRepresentation{} construction} We represent the dental crown mesh $M$ as a 3-channel geometry image $\mathbf{O} \in \mathbb{R}^{H \times W \times 3}$, with the 3 channels encoding the spatial coordinates $(x,y,z)$.
\Cref{fig:geometry_image_representation_path} shows the process via which we obtain the \ourRepresentation{} $\mathbf{O}$ for a given mesh $M$.
Because a tooth is a topological zero-genus surface, we use a global parameterization to encode the tooth with a single chart~\cite{sander2003multi}. 
To construct $\mathbf{O}$ for a given mesh $M$, we first flatten $M$ into a disk $M_{\mathrm{uv}}$ via boundary-fixed harmonic parameterization~\cite{eck_multiresolution_1995}.
We then apply a fisheye-style radial transform to obtain $M_{\mathrm{uv}*}$ so that more space is allocated for the regions closer to the center (corresponding to occlusal surfaces) and less space for the outer rim (containing the axial surfaces). 
To obtain $\mathbf{O}$, we rasterize the 
inverse position map $\pi^{-1}: \mathcal{S} \to \mathbb{R}^3$ onto a regular $H \times W \times 3$ grid. 
The mesh $M^*$ is recovered by triangulating adjacent pixels~\cite{yan_object_2024}.

\begin{figure}[!t]
    \centering
    \includegraphics[trim={0 0 0 0},clip,width=0.95\linewidth]{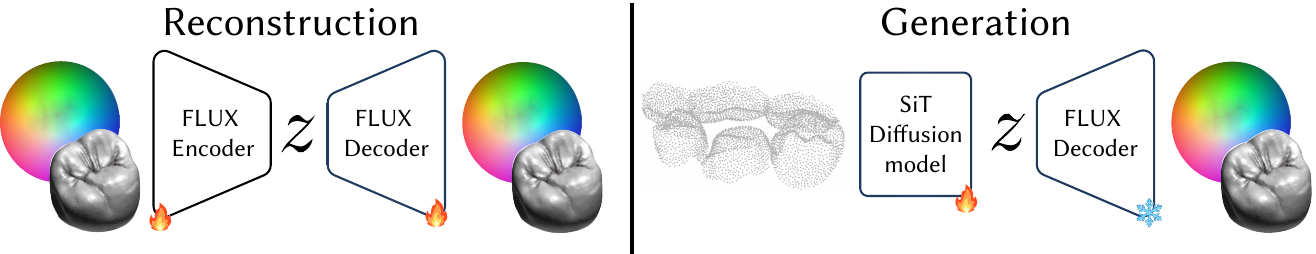}
    \caption{Overview of \ourModel{}. Left: VAE reconstruction from geometry images. Right: conditional generation via SiT diffusion in the learned latent space.
    }
    \label{fig:Geocrown_architecture}
\end{figure}

\smallskip

\mypara{Latent diffusion pipeline} We adopt a latent diffusion framework \cite{rombach_high-resolution_2022} to operate on \ourRepresentation{}s (\cref{fig:Geocrown_architecture}). A variational autoencoder (VAE) encodes the images into latent codes, and a diffusion model generates dental crowns conditioned on the surrounding dentition.

For the VAE, we finetune FLUX \cite{black_forest_labs_flux_2024}. Unlike natural images, \ourRepresentation{}s encode continuous spatial coordinates rather than color values, resulting in a distribution that differs substantially from the data FLUX was pretrained on. Moreover, small latent space errors correspond directly to geometric distortions on the reconstructed surface, making reconstruction fidelity critical. Finetuning the VAE is therefore necessary to faithfully capture geometric variability.\looseness=-1

For the diffusion model, we train a Scalable Interpolant Transformer (SiT) \cite{ma_sit_2024} from scratch. The model is conditioned on two complementary inputs: the prepared tooth's neighbors, the antagonist, and the antagonist's neighbors, and the margin line of the prepared tooth. The margin line delineates the preparation boundary, anchoring the crown's seating position and constraining the generated geometry from below. To retain 3D spatial information, we represent the surrounding dentition as a point cloud sampled from the prepared tooth's neighbors and its antagonist. This context is encoded using a shape encoder \cite{zhang_3dshape2vecset_2023}, trained jointly with the diffusion model. The resulting embedding is integrated via cross-attention at each denoising step, providing spatial and morphological cues from surrounding teeth. This enables the model to infer occlusal surface features from the antagonist and cusp and fissure placement from the neighboring teeth. The decoded \ourRepresentation{} is meshed by triangulating adjacent occupied pixels, i.e. pixels with valid geometric coordinates~\cite{yan_object_2024}.

\smallskip
\mypara{FDI 16 crown dataset} To advance dental crown generation research, we release the \textit{FDI 16 crown dataset}. %
The dataset comprises 11,513 anonymized intraoral scans collected through the 3Shape Dental System,
 with 1,000 scans held out for validation and testing respectively. Each scan is paired with the final crown restoration. %
 We focus on the FDI 16 tooth (upper right first molar) as it is a frequently restored tooth and presents the most complex occlusal morphology of any tooth type, with four to five cusps, an oblique ridge, and a detailed fissure pattern, making it a challenging benchmark. The FDI 16 Tooth Dataset \cite{ye_variational_2023} also includes this tooth, but it only contains isolated natural teeth. Our dataset provides crown restorations paired with surrounding dentition, enabling conditional generation%
. All scans are centered on the FDI 16 crown with the gingiva removed via segmentation using 3Shape Ortho System 2025. 
Where available, the scans include adjacent teeth on the preparation side (FDI 15, 17) and opposing teeth on the antagonist side (FDI 45, 46, 47). We exclude crowns with low overall curvature or significant antagonist intersections, retaining only geometrically complex, well-fitting samples. The crown restorations were designed by dental technicians in routine clinical workflows, reflecting real-world variability in crown quality.\looseness=-1

\section{Experiments}
We evaluate the reconstruction fidelity of \ourModel{} on the FDI 16 Tooth Dataset \cite{ye_variational_2023}, and the generative quality and diversity of the full pipeline on our FDI 16 Crown Dataset through standard metrics and a proposed measure occlusal fit metric to complement them.\looseness=-1

\mypara{Autoencoding using geometry images and comparisons} 
Since the quality of crown generation via \ourModel{} is bounded by the reconstruction fidelity of its autoencoder (AE), we first evaluate our VAE to establish an upper bound on generative performance. In the absence of prior image-based AE baselines for our task, we compare against point cloud-based autoencoders previously benchmarked on the FDI 16 Tooth Dataset \cite{ye_variational_2023}, reporting CD \cite{barrow_parametric_1977} and Earth mover's distances (EMDs) \cite{rubner_earth_2000}. It is worth noting that these point cloud AE baselines all input and output 2,048 points, while our VAE inputs and outputs geometry images. We convert the output \ourRepresentation{} into a mesh, project each GT point onto its surface, and compute CD and EMD between the GT and projected point clouds (denoted \ourModel{}\textsuperscript{$\dagger$}). This point-to-surface evaluation is more favorable than point-to-point comparison, as the continuous mesh surface naturally reduces distances to arbitrary query points.\looseness=-1 

\begin{figure}[!b]
    \centering
    \includegraphics[width=\linewidth]{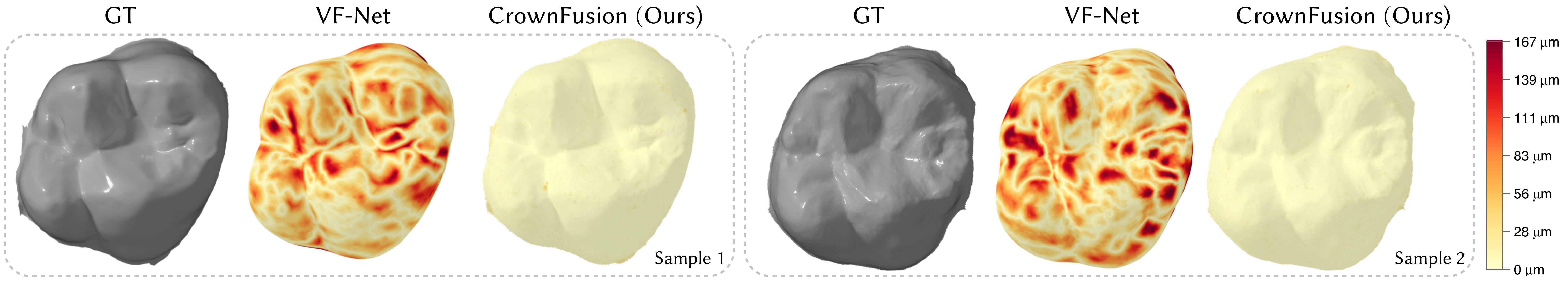}
    \caption{
    Surface distance heatmaps to GT show VF-Net's errors concentrated at cusps and fissures, while our geometry image VAE is close to GT across the crown surface.
    }
    \label{fig:recon_cmp}
\end{figure}

\cref{tab:reconstruction} shows that our VAE yields the lowest error, which can be attributed to the significantly higher spatial resolution of the \ourRepresentation{}s and the FLUX architecture. Visual comparison in Fig.~\ref{fig:recon_cmp} shows that our reconstructions preserve geometric detail beyond what the point-based protocol could capture.\looseness = -1

\begin{table}[t]
	\centering
	\caption{We report reconstruction errors, in CD and EMD, for the FLUX-based VAE in our \ourModel{} and several prior point cloud-based baselines. 
    }
    \label{tab:reconstruction}
    {\fontsize{8}{9}\selectfont
	\begin{tabular}{r rrrr rrr}
		\toprule
		& DPM \cite{luo_diffusion_2021} & SetVAE \cite{kim_setvae_2021} & LION \cite{zeng_lion_2022} & FN \cite{yang_foldingnet_2018} & VF-Net \cite{ye_variational_2023} &CrownFusion\textsuperscript{\textdagger}  \\
		\midrule
		CD ($\times 10^2$) & 10.04 & 21.50 & 5.35 & 5.26  & 1.21 \quad  & \textbf{0.02} \\
		EMD ($\times 10^2$) & 43.98 & 59.24 & 22.85 & 33.67 & 6.30 \quad & \textbf{0.45}\\
		\bottomrule
	\end{tabular}
    }
\end{table}

\mypara{Static occlusal fit}
Next, we train and evaluate the conditional generative pipeline on FDI 16. We compare against DMC \cite{hosseinimanesh_mesh_2025} and VBCD \cite{wei_vbcd_2025}, the two most recent crown completion-based methods with available implementations (\cref{tab:results}). We report the Chamfer distance to a dental crown selected by a dental technician, following prior work~\cite{hosseinimanesh_mesh_2025, yang_dcrownformer_2024, wei_vbcd_2025}. Although our generative approach yields higher reconstruction error than the deterministic VBCD, it outperforms DMC,  despite modeling a full distribution rather than a single output.\looseness=-1

Relying on a single technician-designed crown as ground truth is problematic, since crown designs vary substantially between technicians, and many retain most of the default library tooth morphology with minor adjustments. Similarity to one such reference therefore says little about the clinical quality of a generated crown \cite{mccracken_laboratory_2020, xie_morphological_2025}.
We propose to evaluate occlusal fit by measuring the proximity between the generated occlusal surface and the antagonist dentition. Since teeth rarely align one-to-one, we evaluate occlusal fit against the antagonist tooth and its two nearest neighbors. We sample 30,000 points from these antagonist teeth to enable dense evaluation, and report the mean distance of the closest 10th-percentile of antagonist points to the generated crown surface. This percentile captures the near-contact region most relevant to the occlusal fit. 
We also report intersections between the generated crown and antagonist surfaces, both count and area, as occlusal intersections require significant manual sculpting to correct, unlike proximal intersections which occur on smoother, more accessible surfaces. Minimizing these intersections is therefore essential for reducing manual intervention in clinical workflows.
\begin{table}[!t]
\centering
\caption{
Comparison of \ourModel{} against previous methods for dental crown design. Crown similarity is measured by Chamfer Distance (CD-L2, mm) to a dental technician selected crown. Occlusal fit is evaluated against the three nearest antagonist teeth (by mean distance of 10th-percentile closest points) and surface intersections (count out of 1,000 sampled points, and area in mm\textsuperscript{2}). \ourModel{} achieves the best occlusal fit with fewer intersections and competitive antagonist proximity.   %
}
\label{tab:results}
{
\begin{tabular}{lrrrr}
\toprule
& Crown Similarity $\downarrow$ & \hspace{1mm}Bite Proximity $\downarrow$ & \multicolumn{2}{c}{Intersections $\downarrow$} \\
 \cmidrule(l){2-2} \cmidrule(l){3-3} \cmidrule(l){4-5} 
Model     & CD-L2 (mm) & 10\% closest & \hspace{1mm} num & \hspace{1mm} area (mm\textsuperscript{2})\\
\midrule
DMC~\cite{hosseinimanesh_mesh_2025} & 0.433 & 0.734 & 256 & $1.110$     \\
VBCD~\cite{wei_vbcd_2025} & \textbf{0.353} & $0.747$ & 143 & $2.363$          \\
\ourModel{} (ours)%
& 0.412
& $\mathbf{0.733}$ & \textbf{127} & $\mathbf{0.612}$\\
\midrule
GT & & $0.716$ & $12$ & $0.0437$ \\
\bottomrule
\end{tabular}
}
\end{table}

\begin{figure}[!b]
    \centering
    \includegraphics[width=\linewidth]{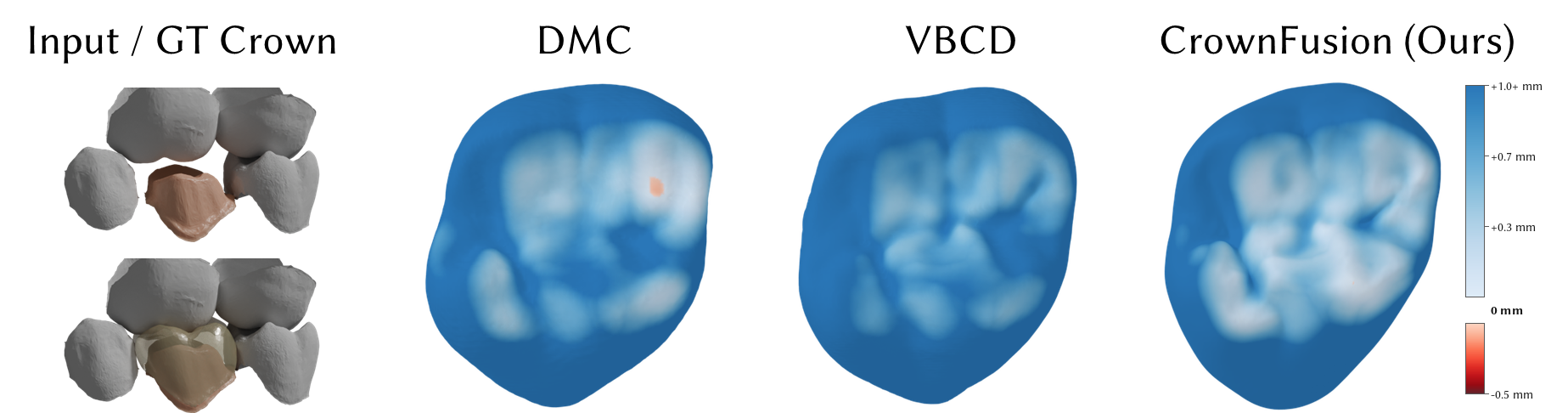}
    \caption{Generation comparison. \textbf{Left:} dental arch with prepared abutment (top) and the reference crown in context (bottom). \textbf{Right:} occlusal views of crowns generated by DMC, VBCD, and \ourModel{}, colored by signed distance to the antagonist (blue: gap; red: intersection). VBCD shows a large uniform gap, indicating poor occlusal contact. DMC exhibits a localized intersection (red). \ourModel{} achieves the closest antagonist proximity with no intersections, and preserves well-defined occlusal anatomy.}
    \label{fig:generation_cmp}
\end{figure}

\begin{figure}[!t]
    \centering
    \includegraphics[width=.95\linewidth]{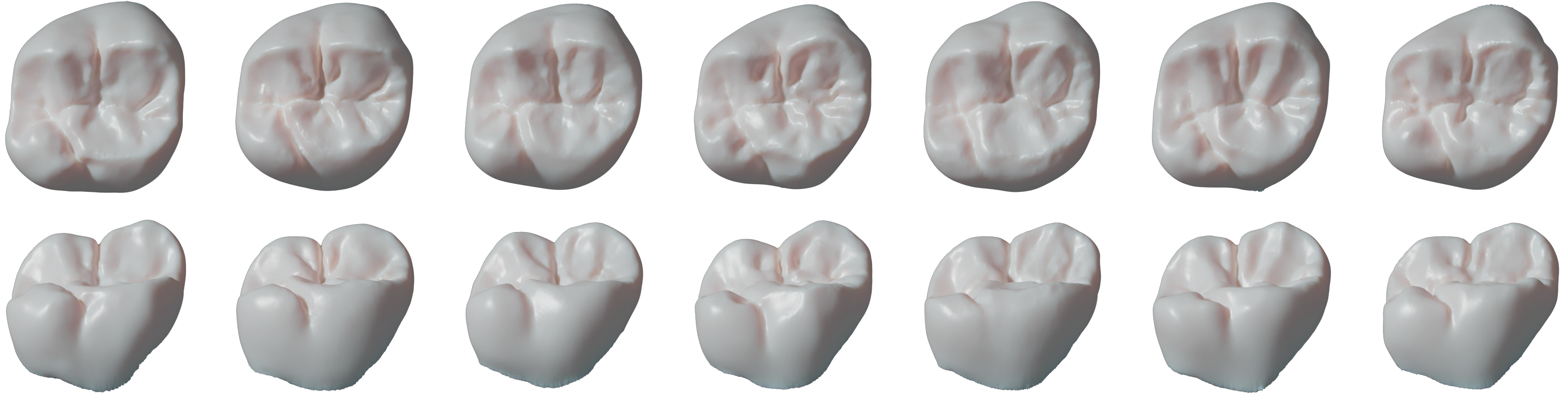}
    \caption{Generated crown samples for the same input. \ourModel{} captures expected axes of variation: fissure depth, secondary fossae detail, and distolingual cusp morphology vary, while marginal ridge height and overall crown outline remain largely stable.
    }
    \label{fig:variants}
\end{figure}

\ourModel{} achieves the closest occlusal alignment to the antagonist teeth (\cref{tab:results}), with fewer intersections than both baselines and roughly half the intersection rate of DMC. When intersections do occur, their area is substantially smaller than in prior methods. \Cref{fig:generation_cmp} shows a representative example.
Ground truth crowns exhibit minimal intersections (12/1000, 0.0437\,mm\textsuperscript{2}), indicating room for future improvement.

\mypara{Expert evaluation}
To complement quantitative metrics, we consulted two dental professionals who assessed generated crowns across all methods. They noted that \ourModel{} produces well-defined occlusal anatomy, rarely generating the  featureless surfaces typical of DMC's overly smooth crowns with shallow fissures and blunt cusps. Across DMC, VBCD, and even the reference crowns, fissures occasionally appear as isolated incisions without corresponding cusp development, yielding anatomically incoherent surfaces; \ourModel{} was less prone to this. It does, however, occasionally produce flattened proximal surfaces that conform too closely to adjacent teeth rather than forming convex contacts, likely reflecting training data where technicians rely on automated fitting tools. This, with occlusal anatomy that does not always match the clinical context, points to conditioning and data curation as areas for improvement.\looseness=-1

\mypara{Morphological variation across samples}
Finally, we examine whether the variation across samples generated for the same patient aligns with the design freedom available in clinical practice. While marginal ridge height and gross cusp placement are constrained by the surrounding dentition, the occlusal surface is the primary locus of design freedom \cite{xie_morphological_2025}, with cusp height, fissure depth, and secondary ridge anatomy varying substantially between technicians.\looseness=-1

Fig.~\ref{fig:variants} shows how the model varies fissure pattern detail, secondary fossae, distolingual cusp, and oblique ridge morphology across samples, consistent with these features carrying more design freedom. Marginal ridge height and overall crown outline remain largely stable, though in isolated cases the variation is larger than clinically ideal. Notably, the features that vary are those most difficult to sculpt manually, as existing CAD tools primarily support local adjustments rather than global morphological changes. Extending the model to offer controllable gross shape diversity through explicit conditioning is a natural next step.\looseness=-1

\mypara{Limitations} Our evaluation focuses on FDI 16; generalization across all tooth types remains to be validated, though the strong generalization of latent diffusion in image synthesis is encouraging. The disk topology of geometry images currently restricts the framework to single-tooth restoration, and conditioning on absent neighbouring teeth is not yet addressed. Generation time ($\sim$5\,min per 8 samples on an RTX 4090) is a practical constraint, though the extensive literature on diffusion sampling acceleration transfers directly to our setting.\looseness=-1

\section{Conclusion}
We presented \emph{\ourModel{}}, the first latent diffusion model for 3D dental crown generation using an image-based teeth representation, 
\emph{\ourRepresentation{}}. %
\ourModel{} generates diverse, well-formed crowns with improved handling of high-curvature dental features, competitive occlusal fit, and fewer antagonist intersections. We release an accompanying dataset of 11,513 annotated intraoral scans and propose an occlusal fit metric based on antagonist proximity.\looseness=-1

\begin{credits}
\subsubsection{\ackname}
This work was partly funded by Innovation Fund Denmark (1044-00172B). SH was supported by the Novo Nordisk Foundation through the Center for Basic Machine Learning Research in Life Science (NNF20OC0062606), VILLUM FONDEN (42062), and the European Research Council (ERC) under the European Union's Horizon programme (grant agreement 101125993). AXC is supported by a CIFAR AI Chair and a NSERC Discovery Grant. HZ is supported by a NSERC Discovery Grant.\looseness=-1

\subsubsection{\discintname}
J.~Z.~Ye and P.~L.~Søndergaard are employees of 3Shape A/S, and the dataset used in this work was collected through 3Shape software. The remaining authors have no competing interests to declare that are relevant to the content of this article.
\end{credits}

\bibliographystyle{splncs04nat}
\bibliography{References}

\end{document}